\documentclass{article}
\usepackage[preprint]{neurips_2024}

\usepackage[utf8]{inputenc}
\usepackage[T1]{fontenc}
\usepackage{times}
\usepackage{inconsolata}
\usepackage{anyfontsize}
\usepackage{microtype}

\usepackage{amsfonts}
\usepackage{amsmath}
\usepackage{amssymb}
\usepackage{mathtools}
\usepackage{nicefrac}
\usepackage{siunitx}
\usepackage{bm}

\usepackage{graphicx}
\usepackage[dvipsnames]{xcolor}

\usepackage{booktabs}
\usepackage{makecell}
\usepackage{multirow}
\usepackage{subcaption}

\usepackage{url}
\usepackage{hyperref}
\usepackage{bookmark}

\usepackage{tikz}
\usetikzlibrary{arrows.meta}
\usetikzlibrary{backgrounds}
\usetikzlibrary{fit}
\usetikzlibrary{math}
\usetikzlibrary{patterns}
\usetikzlibrary{positioning}
\usetikzlibrary{svg.path}

\usepackage{pgfplots}
\usepackage{pgfplotstable}
\pgfplotsset{compat=1.17}
\usepgfplotslibrary{colorbrewer}
\usepgfplotslibrary{colormaps}
\usepgfplotslibrary{groupplots}

\usepackage{natbib}
\setcitestyle{authoryear,open={(},close={)}}

\newcommand{\R}{\mathbb{R}}
\newcommand{\T}{\mathsf{T}}

\newcommand{\vect}[1]{\boldsymbol{#1}}

\newcommand{\tens}[1]{\bm{\mathsf{#1}}}
\newcommand{\norm}[1]{\left\lVert#1\right\rVert}

\DeclareMathOperator{\clamp}{clamp}
\DeclareMathOperator{\sign}{sign}

\DeclareMathOperator{\Attn}{Attn}
\DeclareMathOperator{\FFN}{FFN}
\DeclareMathOperator{\Module}{Module}
\DeclareMathOperator{\Norm}{Norm}
\DeclareMathOperator{\ReLU}{ReLU}

\newcommand{\highlight}[1]{\colorbox{ForestGreen!10}{#1}}
\newcommand{\stt}{\small\ttfamily}

\title{Learning to Skip the Middle Layers of Transformers}
\author{
    Tim Lawson\thanks{Correspondence to \url{tim.lawson@bristol.ac.uk}.}
    \qquad
    Laurence Aitchison\\
    School of Engineering Mathematics and Technology\\
    University of Bristol\\
    Bristol, UK
}

\begin{document}
\maketitle

\begin{abstract}
Conditional computation is a popular strategy to make Transformers more efficient.
Existing methods often target individual modules (e.g., mixture-of-experts layers) or skip layers independently of one another.
However, interpretability research has demonstrated that the middle layers of Transformers exhibit greater redundancy, and that early layers aggregate information into token positions.
Guided by these insights, we propose a novel architecture that dynamically skips a variable number of layers from the \emph{middle outward}.
In particular, a learned gating mechanism determines whether to bypass a symmetric span of central blocks based on the input, and a gated attention mechanism prevents subsequent tokens from attending to skipped token positions.
Residual norms are controlled with a `sandwich' or `peri-layernorm' scheme and gate sparsity with an adaptive regularization loss.
We had aimed to reduce compute requirements for `simpler' tokens and potentially foster an emergent multi-level representational hierarchy but, at the scales investigated, our approach does not achieve improvements in the trade-off between validation cross-entropy and estimated FLOPs compared to dense baselines with fewer layers.
We release our code at \url{https://github.com/tim-lawson/skip-middle}.
\end{abstract}

\section{Introduction}

We want to make Transformers more efficient.
This could be achieved by making individual modules more efficient; for example, many variants on the attention mechanism have been proposed \citep{dong_flex_2024}.
An alternative approach is to reduce the number of parameters activated during inference.
Conditional computation methods decouple a model's capacity (determined by its total parameter count) from its inference cost (determined by the subset of active parameters used for a given input; \citealt{bengio_estimating_2013,bengio_conditional_2016}).
A prominent example is the replacement of feed-forward network (FFN) modules with mixture-of-experts (MoE) layers \citep{shazeer_outrageously_2017}.
These methods reduce computational and memory requirements while enabling parallelization of model components across multiple devices \citep{eigen_learning_2014,lepikhin_gshard_2020,fedus_switch_2022,dai_deepseekmoe_2024}.

One way to reduce the active parameters is to conditionally apply components of a Transformer dependent on the input token.
Then, we can dynamically allocate less compute resources to tokens that are `easier' to process.
Early exiting methods, where deep networks can make predictions at different layers, have a long history in vision \citep{teerapittayanon_branchynet_2016} and language applications \citep{elbayad_depthadaptive_2020,xin_deebert_2020}.
This approach has been used to dynamically skip Transformer layers beyond a certain depth \citep{elhoushi_layerskip_2024,fan_not_2024}.
Other methods skip intermediate components \citep{wang_skipnet_2018}, such as individual modules \citep{csordas_neural_2021,peroni_skip_2024} or entire layers \citep{zeng_learning_2023,raposo_mixtureofdepths_2024}.

We argue that it makes more sense to skip the \emph{middle} layers of Transformers.
Multiple researchers have demonstrated that the middle layers exhibit greater redundancy: for instance, \citet{lad_remarkable_2024} and \citet{gonzalez_leveraging_2025} found that when layers are removed or swapped in pre-trained models, the performance impact is smaller for interventions affecting the more central layers.
This redundancy has been exploited for structured pruning \citep{fan_reducing_2019,gromov_unreasonable_2024,men_shortgpt_2024}.

Interpretability research has also established that early, middle, and late layers in deep networks have different functions.
In language models, early-layer modules convert token-level representations to more natural, semantic features \citep{elhage_softmax_2022,gurnee_finding_2023}.
Furthermore, \citet{kaplan_tokens_2024} have shown that, for multi-token words, the attention mechanism in early Transformer layers aggregates information into the residual vector of the final token in the word.
Conversely, late-layer modules convert semantic features into output tokens: near the output of the network, intermediate states can be decoded to elicit token predictions \citep{nostalgebraist_interpreting_2020,belrose_eliciting_2023}.
Internal activations at the earliest and latest layers are also comparatively distinct from the middle layers through the lens of sparse dictionary learning \citep{lawson_residual_2024}.

Conversions between tokens and semantic features parallel the development of byte-level architectures \citep{xue_byt5_2022,slagle_spacebyte_2024}, which we expect to learn tokenization implicitly.
For example, \citet{pagnoni_byte_2024,neitemeier_hierarchical_2024,kallini_mrt5_2024} instigate a two-layer hierarchy of byte- and token-level representations.
Given \citet{kaplan_tokens_2024}, we might expect that this hierarchy could be profitably extended to levels spanning multiple tokens \citep{ho_block_2024a,videau_bytes_2025}.

Guided by these insights, we propose a gating mechanism that skips a variable number of Transformer blocks from the middle outward, dependent on the input token.
In this way, we can allocate less compute resources to `simpler' inputs by skipping the middle layers, which are more likely to be redundant.
The more central the layer, the fewer tokens it processes, allowing a multi-level hierarchy of representations to emerge.
Unfortunately, at the scales we were able to investigate, this architecture does not improve the trade-off between language-modeling performance and the computational resources required, measured in terms of the estimated FLOPs at inference time were we able to achieve the maximum benefit from the sparsity of the gate values.

\begin{figure}
    \centering
    \begin{tikzpicture}
      [
      arrow/.style={-{Latex[length=1.25mm]}, semithick},
      box/.style={
          draw,
          semithick,
          minimum width=12.5mm,
          inner sep=1mm,
          rounded corners=0.5mm,
          font=\sffamily\scriptsize,
          fill=white
        },
      router/.style={box, fill=Green!25},
      block/.style={box, fill=Orange!25},
      layer/.style={box, draw=none, fill=Gray!10, inner xsep=2.5mm, inner ysep=2mm},
      skip/.style={opacity=0.25},
      label/.style={font=\scriptsize},
      ]
      \tikzmath{
        \xcolsep = 4; \xpathsep = 1; \x0 = 1; \x1 = \x0 + \xcolsep; \x2 = \x1 + \xcolsep;
        \yembed = 0.5;
        \yrouter0 = \yembed + 1; \yblock0 = \yrouter0 + 0.75;
        \yrouter1 = \yblock0 + 1; \yblock1 = \yrouter1 + 0.75;
        \yrouter2 = \yblock1 + 1; \yblock2 = \yrouter2 + 0;
        \yrouter3 = \yblock2 + 1; \yblock3 = \yrouter3 + 0;
        \yhead = \yblock3 + 1;
      }
      \begin{scope}[on background layer]
        \draw[step=1cm, gray, very thin, draw=none] (0, 0) grid (10, 7.5);
      \end{scope}
      \begin{scope}
        \node[box] at (\x0, \yembed) (e00) {Embed};
        \node[box] at (\x0, \yhead) (h00) {Head};
        \node[router] at (\x0, \yrouter0) (r00) {Gate};
        \node[router] at (\x0, \yrouter1) (r01) {Gate};
        \node[] at (\x0, \yrouter2) (r02) {};
        \node[] at (\x0, \yrouter3) (r03) {};
        \node[block] at (\x0, \yblock0) (b00) {Block};
        \node[block] at (\x0, \yblock1) (b01) {Block};
        \node[block] at (\x0, \yblock2) (b02) {Block};
        \node[block] at (\x0, \yblock3) (b03) {Block};
        \draw[arrow] (e00) -- (r00);
        \draw[arrow] (r00) -- (b00);
        \draw[arrow] (b00) -- (r01);
        \draw[arrow] (r01) -- (b01);
        \draw[arrow] (b01) -- (b02);
        \draw[arrow] (b02) -- (b03);
        \draw[arrow] (b03) -- (h00);
        \begin{scope}[on background layer]
          \node[layer, fit={(r00) (b00)}] (l00) {};
          \node[layer, fit={(r01) (b01)}] (l01) {};
          \node[layer, fit={(r02) (b02)}] (l02) {};
          \node[layer, fit={(r03) (b03)}] (l03) {};
        \end{scope}
      \end{scope}
      \begin{scope}
        \node[box] at (\x1, \yembed) (e10) {Embed};
        \node[box] at (\x1, \yhead) (h10) {Head};
        \node[router] at (\x1, \yrouter0) (r10) {Gate};
        \node[router] at (\x1, \yrouter1) (r11) {Gate};
        \node[] at (\x1, \yrouter2) (r12) {};
        \node[] at (\x1, \yrouter3) (r13) {};
        \node[block] at (\x1, \yblock0) (b10) {Block};
        \node[block, skip] at (\x1, \yblock1) (b11) {Block};
        \node[block, skip] at (\x1, \yblock2) (b12) {Block};
        \node[block] at (\x1, \yblock3) (b13) {Block};
        \draw[arrow] (e10) -- (r10);
        \draw[arrow] (r10) -- (b10);
        \draw[arrow] (b10) -- (r11);
        \draw[arrow] (b13) -- (h10);
        \begin{scope}[on background layer]
          \node[layer, fit={(r10) (b10)}] (l10) {};
          \node[layer, fit={(r11) (b11)}] (l11) {};
          \node[layer, fit={(r12) (b12)}, skip] (l12) {};
          \node[layer, fit={(r13) (b13)}] (l13) {};
        \end{scope}
        \draw[arrow] (r11)
        -- (\x1 + \xpathsep, \yrouter1)
        -- (\x1 + \xpathsep, \yrouter3 - 0.5)
        -- (\x1, \yrouter3 - 0.5)
        -- (b13);
      \end{scope}
      \begin{scope}
        \node[box] at (\x2, \yembed) (e20) {Embed};
        \node[box] at (\x2, \yhead) (h20) {Head};
        \node[router] at (\x2, \yrouter0) (r20) {Gate};
        \node[router, skip] at (\x2, \yrouter1) (r21) {Gate};
        \node[] at (\x2, \yrouter2) (r22) {};
        \node[] at (\x2, \yrouter3) (r23) {};
        \node[block, skip] at (\x2, \yblock0) (b20) {Block};
        \node[block, skip] at (\x2, \yblock1) (b21) {Block};
        \node[block, skip] at (\x2, \yblock2) (b22) {Block};
        \node[block, skip] at (\x2, \yblock3) (b23) {Block};
        \draw[arrow] (e20) -- (r20);
        \begin{scope}[on background layer]
          \node[layer, fit={(r20) (b20)}] (l20) {};
          \node[layer, fit={(r21) (b21)}, skip] (l21) {};
          \node[layer, fit={(r22) (b22)}, skip] (l22) {};
          \node[layer, fit={(r23) (b23)}, skip] (l23) {};
        \end{scope}
        \draw[arrow] (r20)
        -- (\x2 + \xpathsep, \yrouter0)
        -- (\x2 + \xpathsep, \yhead0 - 0.5)
        -- (\x2, \yhead - 0.5)
        -- (h20);
      \end{scope}
      \node[label] at (\x0 - 1.625, 0.5 * \yrouter0 + 0.5 * \yblock0) {\sffamily Layer 0};
      \node[label] at (\x0 - 1.625, 0.5 * \yrouter1 + 0.5 * \yblock1) {\sffamily Layer 1};
      \node[label] at (\x0 - 1.625, \yblock2) {\sffamily Layer 2};
      \node[label] at (\x0 - 1.625, \yblock3) {\sffamily Layer 3};
      \node[label] at (\x0 + 1.25, \yembed) {\sffamily Token 0};
      \node[label] at (\x1 + 1.25, \yembed) {\sffamily Token 1};
      \node[label] at (\x2 + 1.25, \yembed) {\sffamily Token 2};
      \node[label] at (\x0 + 1.625, \yrouter0 - 0.15) {$g^{(0,0)} > 0$};
      \node[label] at (\x0 + 1.625, \yrouter1 - 0.15) {$g^{(0,1)} > 0$};
      \node[label] at (\x1 + 1.625, \yrouter0 - 0.15) {$g^{(1,0)} > 0$};
      \node[label] at (\x1 + 1.625, \yrouter1 - 0.15) {$g^{(1,1)} = 0$};
      \node[label] at (\x2 + 1.625, \yrouter0 - 0.15) {$g^{(2,0)} = 0$};
      \node[font=\sffamily\scriptsize, align=center] at (\x1 + 1.8, \yblock1 + 0.5) {Skip blocks\\ 1–2};
      \node[font=\sffamily\scriptsize, align=center] at (\x2 + 1.8, \yblock1 + 0.5) {Skip blocks\\ 0–3};
    \end{tikzpicture}
    \caption{An illustration of our proposed architecture (with four layers or blocks). We compute a scalar gate value for each token position and block in the first half of the model. If the gate at block $\ell$ is zero, we skip the Transformer blocks between $\ell$ and $L-\ell$ for the token, and prevent other tokens from attending to its position in the corresponding self-attention modules.}
    \label{fig:diagram}
\end{figure}
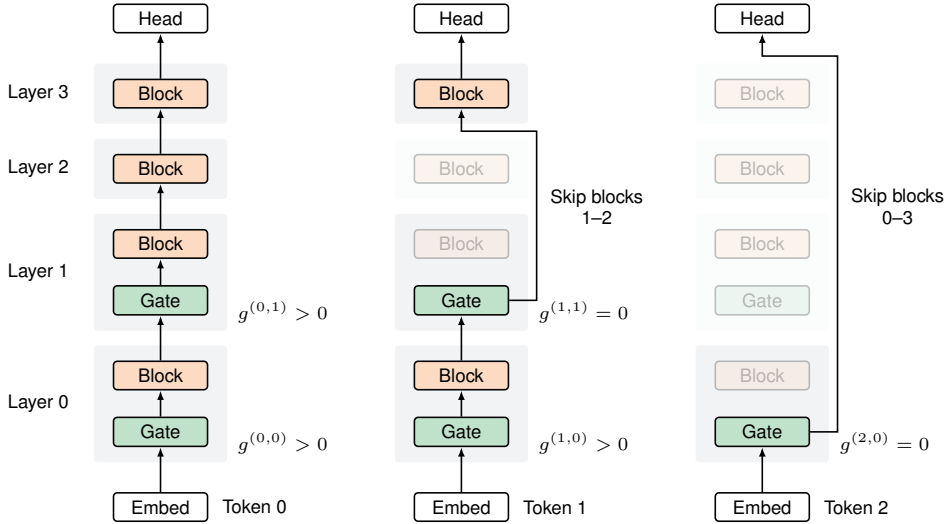

\section{Model architecture}

A standard decoder-only Transformer has $L$ layers or blocks, each of which comprises a self-attention and FFN module.
We denote the input to layer $\ell$ at token position $i$ by $\vect{h}^{(i,\ell)} \in \R^d$ where $d$ is the model dimension, and the inputs at all token positions $i \in 1..N$ by $\tens{H}^{(\ell)} \in \R^{N \times D}$.

At a high level, the standard Transformer is given by (omitting layer normalization):
\begin{align*}
    \label{eqn:standard_transformer}
    \vect{h}^{(i,0)}    & = \operatorname{Embed}(i) \\
    \vect{a}^{(i,\ell)} & = \vect{h}^{(i,\ell-1)} + \Attn(\tens{H}^{(\ell-1)}) \\
    \vect{h}^{(i,\ell)} & = \vect{a}^{(i,\ell)} + \FFN(\vect{a}^{(i,\ell)}) \\
    \vect{y}^{(i)}      & = \operatorname{Head}(\vect{h}^{(i,L)}).
\end{align*}
We propose to modify this architecture like so:
\begin{align*}
    \vect{a}^{(i,\ell)} & = \vect{h}^{(i,\ell-1)} +
        \highlight{$g^{(i,\ell)}
        \operatorname{GatedAttn}$}(\tens{H}^{(\ell-1)}, \highlight{$\vect{g}^{(\ell)}$}) \\
    \vect{h}^{(i,\ell)} & = \vect{a}^{(i,\ell)} +
        \highlight{$g^{(i,\ell)}$}
        \FFN(\vect{a}^{(i,\ell)}). 
\end{align*}
The modifications are \highlight{highlighted}.
If the gate value $g^{(i,\ell)} \in \R$ for a token position $i$ is zero, then we do not need to compute the corresponding outputs of the attention and feed-forward modules.
The gating mechanism thus functions as a kind of router (Figure~\ref{fig:diagram}).

\subsection{Gating mechanism}
\label{sec:gating_mechanism}

For each block $\ell < L/2$ in the first half of the model, we introduce a linear layer that outputs a scalar \textbf{\underline{s}oft mask} value $s^{(i,\ell)} \geq 0$, which accumulates over the first half of the model:
\begin{equation}
    \label{eqn:soft_mask}
    s^{(i,\ell)} = \ReLU\left(\vect{w}^{(\ell)} \cdot \vect{h}^{(i,\ell)} + b^{(\ell)}\right)
    ,\quad
    S^{(i,\ell)} = \sum_{\ell' \leq \ell} s^{(i,\ell')}.
\end{equation}
When the accumulated soft mask value $S^{(i,\ell)} \geq 1$, we skip processing the residual vector at token position $i$ by the Transformer blocks $[\ell, L-\ell)$.
Hence, for each block $\ell \geq L/2$ in the second half of the model, we use the accumulated soft mask value $S^{(i, L-\ell-1)}$ of the opposing block.

The corresponding scalar \textbf{\underline{g}ate} value $g^{(i,\ell)} \in [0,1]$ is:
\begin{equation}
    \label{eqn:gate}
    g^{(i,\ell)} = \begin{cases}
        1 - \clamp\left(S^{(i, \ell)},     0, 1\right)
        & \text{if } \ell < L/2 \\[0.5ex]
        1 - \clamp\left(S^{(i, L-\ell-1)}, 0, 1\right)
        & \text{if } \ell \geq L/2
    \end{cases}
\end{equation}

The sparsity of the gate values $g^{(i,\ell)}$ determines the reduction in the number of active parameters: for a single token $i$, it is the number of blocks at which the gate value is zero multiplied by the number of parameters in a Transformer block $N_B$.
With multiple tokens, the reduction is $2N_B \sum_{\ell < L/2} z_\ell$, where $z_\ell$ is the sparsity of the gate values before tokens are processed by block $\ell$.

The gating mechanism introduces $(d+1)L/2$ parameters $\vect{w}^{(\ell)}$ and $b^{(\ell)}$, i.e., $d+1$ for each block $\ell < L/2$ in the first half of the model.
If all these parameters are zero, all the gate values equal one and we exactly recover the equivalent dense Transformer (which forms the baselines in Section~\ref{sec:results}).

\subsection{Gated attention}
\label{sec:gated_attention}

We also prevent other tokens from attending to gated tokens in attention modules.
The $\operatorname{GatedAttn}$ module we incorporate modifies the attention mechanism such that, when the gate value for a token is zero, subsequent tokens do not attend to the gated token:
\begin{align}
    \vect{o}_i = \frac{\sum_{j<i} \highlight{$g_j$} \exp(\vect{q}_i^\T \vect{k}_j) \vect{v}_j}{\sum_{j<i} \highlight{$g_j$} \exp(\vect{q}_i^\T \vect{k}_j)}
\end{align}
This is equivalent to adding $\ln g_j$ to the pre-softmax attention logits, and can be implemented straightforwardly as a score modification within the FlexAttention framework \citep{dong_flex_2024}.
In practice, we apply a lower bound of $\epsilon = \num{1e-6}$ to $g_j$ before $\ln$ to prevent infinities.

Our attention mechanism is similar to the `Forgetting Attention' proposed by \citet{lin_forgetting_2024}, except that we compute a single gate value that applies to every attention head, whereas they compute a gate value for each attention head.
We require a single gate value to decide whether to prevent an entire Transformer block from processing the token.

\subsection{Layer normalization}
\label{sec:layer_normalization}

Modern Transformers typically use the `pre-layernorm' scheme (pre-LN), where layer normalization operations are applied to the residual inputs to the attention and FFN modules:
\begin{equation*}
    \vect{y} = \vect{x} + \Module(\Norm(\vect{x})).
\end{equation*}
With this scheme, the norms of residual activation vectors grow with depth and later modules produce outputs with greater norms \citep{lawson_residual_2024,csordas_moeut_2024a,kim_periln_2025}.

The gating mechanism we propose effectively introduces skip connections between opposing pairs of Transformer blocks (Section~\ref{sec:gating_mechanism}).
Like \citet{csordas_moeut_2024a}, who consider a Universal Transformer (UT) with a single, shared block, we want later modules to accept the outputs of early and late blocks.
We address this problem by using the `sandwich' LN scheme proposed in \citet{ding_cogview_2021}, called `peri-layernorm' by \citet{kim_periln_2025}, where layer normalization operations are applied to both the residual input to \emph{and} output of the attention and FFN modules:
\begin{equation*}
    \vect{y} = \vect{x} + \Norm(\Module(\Norm(\vect{x}))).
\end{equation*}
This scheme differs from the `peri-layernorm' of \citet{csordas_moeut_2024a}, who apply a layer normalization operation ``around (but not on) the residual connections.''

\subsection{Controlling sparsity}
\label{sec:controlling_sparsity}

Our gated architecture reduces the number of parameters activated during the forward pass proportional to the fraction of gate values that are exactly zero, i.e., the mean gate sparsity (Section~\ref{sec:gating_mechanism}).
With only the standard cross-entropy loss, optimization tends to activate more parameters to improve performance, so we need to control the sparsity of the gate values.

We achieve this by introducing a regularization loss based on the mean and variance of the gate values, with adaptive coefficients updated proportional to the deviations of the mean and variance from layer-wise targets.
The mean term incentivizes smaller gate values; the variance term incentivizes a non-uniform distribution such that some (but not all) gate values are zero.

We denote the mean and variance of the gate values over the token positions in a batch by:
\begin{equation}
    \overline{g}_\ell = \frac{1}{N} \sum_{i=1}^{N} g^{(i,\ell)}
    \ ,\quad
    s^2_\ell = \frac{1}{N} \sum_{i=1}^{N} \left( g^{(i,\ell)} - \overline{g}_\ell \right)^2.
\end{equation}
The targets at layer $\ell$ for the population mean and variance of the gate values over token positions are $\mu^*_\ell$ and $\sigma^{2*}_\ell$, respectively.
Except where noted, we choose the mean targets $\mu^*_\ell$ as linearly spaced values between an initial target $\mu^*_0$ and a final target $\mu^*_{L/2}$.
We choose the variance targets $\sigma^{2*}_\ell$ as the variance of the Bernoulli distribution with $p = \mu^*_\ell$, i.e., $\sigma^{2*}_\ell = \mu^*_\ell (1 - \mu^*_\ell)$.

\begin{table}[]
  \centering
  \renewcommand{\arraystretch}{1.25}
  \begin{tabular}[t]{lll}
    \toprule
    \textbf{Name} & \textbf{Loss} & \textbf{Update rule} \\
    \midrule
    \stt sparsity & $\frac{1}{L} \sum_{\ell=1}^{L} \alpha_\ell\, \overline{g}_\ell$ & - \\
    \midrule
    \stt sparsity\_variance & \multirow{3}{*}{$\frac{1}{L} \sum_{\ell=1}^{L} \left( \alpha_\ell\, \overline{g}_\ell + \beta_\ell\, s^2_\ell \right)$} & - \\
    \stt adaptive & & $\alpha_{i+1} = \alpha_i + \gamma\,\sign(\overline{g}_\ell - \mu^*_\ell)$ \\
    \stt proportional & & $\alpha_{i+1} = \alpha_i + \gamma\,(\overline{g}_\ell - \mu^*_\ell)$ \\
    \midrule
    \stt sparsity\_variance\_l2 & \multicolumn{2}{l}{$\frac{1}{L} \sum_{\ell=1}^{L} \Bigl( \alpha_\ell\, \norm{\overline{g}_\ell - \mu^*_\ell}_2^2 \ +\ \beta_\ell\, \norm{s^2_\ell - \sigma^{2*}_\ell}_2^2 \Bigr)$} \\
    \bottomrule
  \end{tabular}
  \par\bigskip
  \caption{Alternative techniques to control the sparsity of the gate values. Recall that $\overline{g}_\ell$ and $s^2_\ell$ are the mean and variance of the gate values over the token positions in a batch, and $\mu^*_\ell$ and $\sigma^{2*}_\ell$ are the targets at layer $\ell$ for the mean and variance of the gate values, respectively. For the techniques with adaptive coefficients, $\alpha_\ell$ and $\beta_\ell$ are updated by the same algorithm.}
  \label{tab:controllers}
\end{table}

We denote the adaptive coefficients for the mean and variance of the gate values at layer $\ell$ by $\alpha_\ell$ and $\beta_\ell$, respectively, and initialize the coefficients to zero.
The regularization loss is then:
\begin{equation}
    \mathcal{L} = \frac{1}{L} \sum_{\ell=1}^{L} \left(
        \alpha_\ell\, \overline{g}_\ell + \beta_\ell\, s^2_\ell
    \right).
\end{equation}
After every training step, we update each coefficient by the following rule:
\begin{equation}
    \alpha_{\ell,i+1} =
    \begin{cases}
        \alpha_{\ell,i} + \gamma \left( \overline{g}_\ell - \mu^*_\ell \right)
        & \text{if} \left( \overline{g}_\ell - \mu^*_\ell \right) > \delta
        \\[0.5ex]
        \alpha_{\ell,i}
        & \text{otherwise}
    \end{cases}
\end{equation}
The updates are thus proportional to the differences from the target values.
We choose the update multiplier $\gamma = \num{1e-3}$ and tolerance $\delta = \num{1e-2}$ based on observations in small-scale experiments.
We explored alternative control mechanisms (Table~\ref{tab:controllers}), but these performed worse empirically.

\section{Results}
\label{sec:results}

\begin{figure}
  \centering
  \begin{tikzpicture}
    \begin{groupplot}[
        scale only axis,
        width=2.25in,
        height=2.5in,
        group style={
            group size=2 by 1,
            horizontal sep=0.25in,
            y descriptions at=edge left,
          },
        ymin=3.15,
        ymax=3.65,
        ylabel={Cross-entropy},
        y label style={anchor=south,at={(-0.15,.5)}},
        grid=major,
        tick pos=left,
        xtick align=outside,
        ytick align=outside,
        major tick length=3pt,
        legend cell align={left},
        legend columns=1,
        legend style={
            anchor=north east,
            column sep=0.1in,
            draw=none,
            font=\small,
          },
        cycle list/Paired,
        cycle list shift=1,
        every node near coord/.append style={xshift=0pt, yshift=0pt, font=\footnotesize},
      ]
      \nextgroupplot[xlabel={Inference FLOPs}]
      \addplot+ [only marks, mark=*, mark size=1.5pt] table [col sep=comma, x={infer_flops}, y={loss}] {figures/flops_loss_fineweb-baseline.csv};
      \addplot+ [only marks, mark=x, mark size=2.5pt, thick] table [col sep=comma, x={infer_flops}, y={loss}] {figures/flops_loss_fineweb-nocontrol.csv};
      \addplot+ [only marks, mark=+, mark size=2.5pt, thick] table [col sep=comma, x={infer_flops}, y={loss}] {figures/flops_loss_fineweb-gated.csv};
      \nextgroupplot[xlabel={Sparsity}, enlarge x limits=0.15, xticklabel style={/pgf/number format/fixed, /pgf/number format/precision=2}]
      \addplot+ [only marks, mark=*, mark size=1.5pt, point meta=explicit symbolic, nodes near coords, nodes near coords align={anchor=east}] table [col sep=comma, x={zeros}, y={loss}, meta={model.n_layers}] {figures/flops_loss_fineweb-baseline.csv};
      \addplot+ [only marks, mark=x, mark size=2.5pt, thick, point meta=explicit symbolic] table [col sep=comma, x={zeros}, y={loss}, meta={model.n_layers}] {figures/flops_loss_fineweb-nocontrol.csv};
      \addplot+ [only marks, mark=+, mark size=2.5pt, thick] table [col sep=comma, x={zeros}, y={loss}] {figures/flops_loss_fineweb-gated.csv};
      \addlegendentry{Dense baseline}
      \addlegendentry{Gated (without control)}
      \addlegendentry{Gated (with control)}
    \end{groupplot}
  \end{tikzpicture}
  \caption{Performance comparisons between our gated Transformer architecture and baseline models with between 2 and 12 layers (labeled). All gated models with control are variants of the 12-layer architecture. We measured cross-entropy over 100M tokens from the FineWeb validation set. The estimated FLOPs for a single forward pass (left) assume that the maximum computational benefit is achieved from the final sparsity of the gate values over the validation set (right).}
  \label{fig:results}
\end{figure}
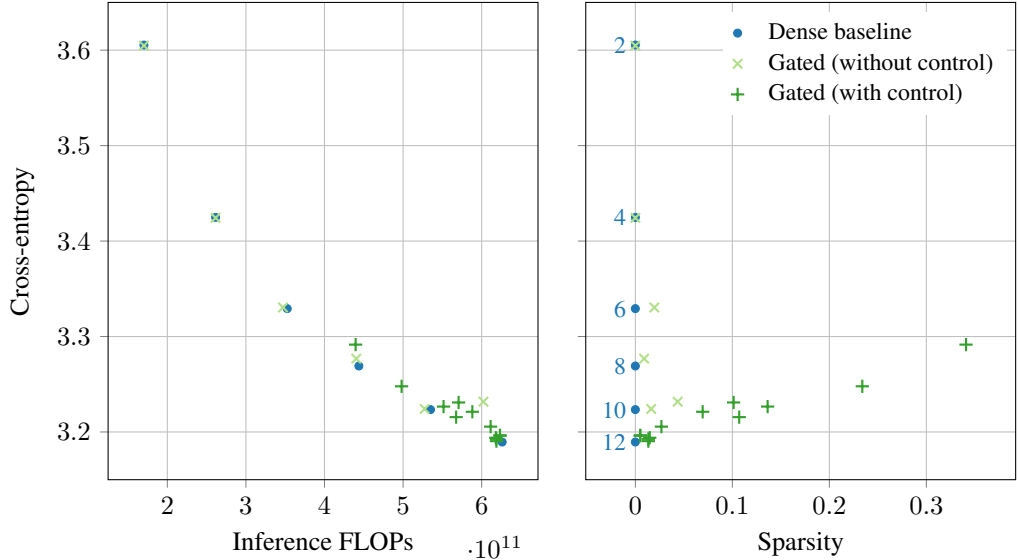

We evaluated the performance of our gated Transformer architecture in terms of the validation cross-entropy when pre-trained on the FineWeb dataset \citep{penedo_fineweb_2024}.
As baselines, we trained equivalent dense models with no gating mechanism and between 2 and 12 layers.
We measured the compute requirements of each model in terms of the estimated floating-point operations (FLOPs) for a single forward pass (batch size 1), \emph{assuming} that we were able to achieve the maximum possible benefit due to the sparsity of the gate values.
The actual compute requirements of the gated and dense models are similar.
Except where noted, we used the hyperparameters in Table~\ref{tab:hyperparameters_fineweb}.

Figure~\ref{fig:results} shows that, without controlling the sparsity of the gates, the mean sparsity tends toward zero and gated models perform similarly to dense baselines (right).
With the initial target for the mean gates $\mu^*_0$ fixed to 1, as we decrease the final target $\mu^*_{L/2}$ from 1 to 0, the sparsity increases and the estimated FLOPs decrease (left).
However, the proposed architecture does not improve the cross-entropy over dense baselines with fewer layers.

\subsection{Experimental details}

We trained all models on a randomly-sampled subset of the FineWeb dataset with approximately 10B tokens \citep{penedo_fineweb_2024}, pre-tokenized with the GPT-2 tokenizer via the TikToken library \citep{radford_language_2019,ouyang_training_2022}.
The validation set contained approximately 100M tokens.
We used a global batch size of 512 sequences (\num{524288} tokens) with data parallelism and gradient accumulation over a per-device batch size of 32 sequences on 4 NVIDIA A100 or GH200 GPUs.

We based the underlying Transformer models on the reference implementation of Llama 3  \citep{grattafiori_llama_2024}.
In particular, we used:
Grouped Query Attention (GQA; \citealt{ainslie_gqa_2023});
Rotary Positional Embeddings (RoPE; \citealt{su_roformer_2024});
Gated Linear Unit FFNs with Swish activation (SwiGLU; \citealt{shazeer_glu_2020}); and
Root Mean Square (RMSNorm) layer normalization \citep{zhang_root_2019}.
The key difference relative to Llama 3 is that we used the Sandwich-LN scheme \citep{ding_cogview_2021,kim_periln_2025} instead of Pre-LN.
We initialized RMSNorm parameters to one and sampled all others from the normal distribution with mean zero and standard deviation \num{0.02}.

\begin{table}
  \centering
  \begin{subtable}[t]{0.425\textwidth}
    \centering
    \begin{tabular}[t]{lr}
      \toprule
      \textbf{Parameter}              & \textbf{Value} \\
      \midrule
      \multicolumn{2}{l}{\stt model}                   \\
      \quad \stt dim                  & \num{768}      \\
      \quad \stt n\_layers            & \num{12}       \\
      \quad \stt n\_heads             & \num{12}       \\
      \quad \stt n\_kv\_heads         & \num{12}       \\
      \quad \stt vocab\_size          & \num{50257}    \\
      \quad \stt ffn\_dim\_multiplier & \num{4}        \\
      \quad \stt multiple\_of         & \num{256}      \\
      \quad \stt norm\_eps            & \num{1e-5}     \\
      \quad \stt rope\_theta          & \num{10000}    \\
      \quad \stt use\_scaled\_rope    & \stt False     \\
      \quad \stt max\_seq\_len        & \num{1024}     \\
      \quad \stt initializer\_range   & \num{0.02}     \\
      \bottomrule
    \end{tabular}
  \end{subtable}
  \begin{subtable}[t]{0.425\textwidth}
    \centering
    \begin{tabular}[t]{lr}
      \toprule
      \textbf{Parameter}       & \textbf{Value} \\
      \midrule
      \multicolumn{2}{l}{\stt data}                    \\
      \quad \stt batch\_size          & \num{512}      \\
      \quad \stt device\_batch\_size  & \num{32}       \\
      \midrule
      \multicolumn{2}{l}{\stt optimizer}        \\
      \quad \stt lr            & \num{0.001}    \\
      \quad \stt beta1         & \num{0.8}      \\
      \quad \stt beta2         & \num{0.95}     \\
      \quad \stt eps           & \num{1e-10}    \\
      \quad \stt weight\_decay & \num{0}        \\
      \midrule
      \multicolumn{2}{l}{\stt scheduler}        \\
      \quad \stt warmup\_steps & \num{0.1}      \\
      \quad \stt start\_factor & \num{0.1}      \\
      \bottomrule
    \end{tabular}
  \end{subtable}
  \par\bigskip
  \caption{Default hyperparameters.
    The Transformer model architecture is based on Llama 3 \citep{grattafiori_llama_2024} with similar dimensions to GPT-2 small \citep{radford_language_2019}; we used the AdamW optimizer \citep{loshchilov_decoupled_2019} with linear warm-up and cosine decay.}
  \label{tab:hyperparameters_fineweb}
\end{table}

The training codebase is based on the `nanoGPT speedrun' repository \citep{karpathy_karpathy_2025,jordan_kellerjordan_2025}.
We used the AdamW optimizer with a single learning rate for all model parameters \citep{kingma_adam_2017,loshchilov_decoupled_2019}, and a two-stage learning-rate scheduler with linear warm-up over 10\% of the training steps, starting at 10\% of the maximum learning rate, and cosine decay over the remaining steps.
Lastly, we performed forward passes in \texttt{bfloat16} with automatic mixed precision in PyTorch (except manually converting attention logits to \texttt{float32}).

\section{Related work}
\label{sec:related_work}


Conditional computation decouples a model's total parameter count from its inference cost by activating only a subset of parameters for a given input \citep{bengio_estimating_2013,eigen_learning_2014,bengio_conditional_2016}.
A prominent application of this principle is the use of Mixture-of-Experts layers, which replace FFN modules with a larger set of `expert' sub-networks, of which only a few are selected by a router to process each input \citep{shazeer_outrageously_2017,lepikhin_gshard_2020,fedus_switch_2022,dai_deepseekmoe_2024}.
While related methods like \citet{zhang_mixture_2022,csordas_switchhead_2024,jin_moh_2024} can be effective, they operate on individual modules (e.g., FFNs or attention) and often use routing strategies that enforce a fixed computational budget per token \citep[cf.][]{wang_remoe_2024}.
Our approach differs by applying conditional computation to entire Transformer blocks, and dynamically allocating a variable number of blocks to each token based on its processing needs, a strategy that is also compatible with module-level techniques.


Another line of work to improve efficiency is dynamically altering the network depth.
Early exiting methods allow a model to generate predictions at intermediate layers, halting computation for `easy' inputs \citep{teerapittayanon_branchynet_2016,elbayad_depthadaptive_2020,xin_deebert_2020}.
More recent variants of this approach dynamically skip all layers beyond a certain depth \citep{elhoushi_layerskip_2024,fan_not_2024}.
In contrast, our method is motivated by empirical findings that the middle layers of Transformers exhibit greater redundancy \citep{lad_remarkable_2024,gonzalez_leveraging_2025}.
These findings have been leveraged by structured pruning, which removes layers statically after training \citep{fan_reducing_2019,gromov_unreasonable_2024,men_shortgpt_2024}.
Our work is distinct in that it targets the more redundant middle layers for skipping and does so dynamically during inference.


Several methods have explored skipping entire Transformer blocks.
For instance, the copy gate proposed by \citet{csordas_neural_2021} modulates a block's contribution, but still requires the full computation of the block's output.
Our gating mechanism, however, ensures that computation for skipped blocks could be avoided entirely.
Mixture-of-Depths (MoD) models \citep{raposo_mixtureofdepths_2024} process only the top-$k$ tokens at each block, enforcing a fixed computational budget for the sequence.
Our method is different because it determines the computational depth for each token individually, allowing it to adapt to token-specific complexity rather than a sequence-wide budget.
Approaches like SkipNet \citep{wang_skipnet_2018} also enable layer skipping but are not specifically designed for the unique redundancy patterns observed in the middle layers of Transformers.


Our gated attention mechanism prevents attention to tokens that have been masked out, which is functionally similar to the `Forgetting Attention' proposed by \citet{lin_forgetting_2024}.
However, their approach computes separate gates for each attention head, whereas our approach uses a single gate per token to decide whether to skip an entire Transformer block, which is essential for the block-level computational savings we target.

Finally, our work relates to hierarchical Transformers that process sequences at multiple predefined levels (e.g., byte- and token-level; \citealt{pagnoni_byte_2024,neitemeier_hierarchical_2024,kallini_mrt5_2024}).
Our middle-outward skipping approach offers the potential for a more flexible, emergent hierarchy, where deeper layers process a dynamically determined subset of more complex representations.

\section{Conclusion}

We introduced a novel Transformer architecture that dynamically skips a variable number of middle layers, guided by interpretability research suggesting these layers are the most redundant.
The mechanism uses a learned gate to bypass a symmetric span of central blocks based on the input token, with the goals of reducing compute for simpler tokens and allowing a representational hierarchy to emerge.
Our experiments showed that, at small scales, this architecture did not yield improvements in the trade-off between validation performance and estimated inference FLOPs when compared to simply training dense baseline models with fewer layers.
The anticipated benefits of this architectural prior may only become apparent at significantly larger model scales, where the redundancy of middle layers is more pronounced and the relative overhead of the gating mechanism is smaller.
Despite these results, we believe the principle of using insights from model internals to design more efficient and structured architectures remains a valuable direction for future research.

\bibliographystyle{plainnat}
\bibliography{main}

\end{document}